\documentclass{bmvc2k}

\usepackage{booktabs}
\usepackage{graphicx}
\usepackage{multirow}
\usepackage{amssymb}
\usepackage[table]{xcolor}
\usepackage{capt-of}
\usepackage{pifont}
\usepackage{marvosym}
\newcommand{\cmark}{\ding{51}}
\newcommand{\xmark}{\ding{55}}
\newcommand\blfootnote[1]{%
  \begingroup
  \renewcommand\thefootnote{}%
  \footnotetext{#1}%
  \endgroup
}

\title{Multimodal Action Diffusion for Robust End-to-End Autonomous Driving}

\addauthor{Jorge Daniel Rodríguez-Vidal\textsuperscript{\Letter,}}{jdrodriguez@cvc.uab.cat}{1}
\addauthor{Diego Porres}{dporres@cvc.uab.cat}{1}
\addauthor{Gabriel Villalonga Pineda}{gvillalonga@cvc.uab.cat}{1,2}
\addauthor{Antonio M. López Peña}{antonio@cvc.uab.es}{1,2}

\addinstitution{
 Computer Vision Center (CVC)\\
 Barcelona, Spain
}
\addinstitution{
 Universitat Autònoma de Barcelona (UAB)\\
 Barcelona, Spain
}

\runninghead{Rodríguez-Vidal \bmvaEtAl}{Multimodal E2E Autonomous Driving}

\begin{document}

\maketitle

\blfootnote{\textsuperscript{\Letter} Corresponding author.}

\begin{abstract}
End-to-End Autonomous Driving (E2E-AD) systems have largely converged on predicting intermediate trajectory waypoints, delegating final control to hand-crafted controllers with GPS access. Direct control-signal prediction (outputting throttle, steer and brake in an end-to-end fashion) remains underexplored, and critically, the role of \emph{action multimodality} in such systems is not well understood. We argue that moving beyond deterministic, single-action outputs is not merely a modelling choice, but a key driver of driving performance, representational quality, and training stability. To validate this, we introduce the Action Diffusion Transformer (ADT), an anchor-free diffusion transformer trained with a MSE objective that natively models the multimodal distribution of plausible driving actions. Rather than committing to a single deterministic command, ADT generates $K$ action candidates and selects the most suitable one at inference via Nearest Neighbour Matching (NNM). Beyond strong benchmark numbers, we show that action multimodality yields measurable benefits in learned representations and behavioral consistency — effects that deterministic architectures cannot replicate. ADT surpasses previous state-of-the-art on the challenging closed-loop Bench2Drive benchmark while achieving $10\times$ lower latency, demonstrating that expressive, multimodal action modelling is both practically efficient and conceptually essential for robust end-to-end driving. Project page: \url{https://jorgedanielrodrividal.github.io/multimodal-e2e-autonomous-driving/}
\end{abstract}

\section{Introduction}
\label{sec:intro}

End-to-End Autonomous Driving (E2E-AD)~\cite{Chen24E2EAD} has benefited from recent advances in generative modelling ~\cite{Ho20DDPM, Karras22EDM, Chi23DiffusionPolicy} and simulation ~\cite{Dosovitskiy17, Dauner24NAVSIM, Franchi22MUAD}. In particular, diffusion models offer a natural way to represent the inherently multi-modal nature by enabling the generation of diverse plausible predictions. However, despite this probabilistic capacity, most E2E-AD models remain deterministic as they collapse the learned distribution into a single waypoint trajectory or control output.

\begin{figure*}[t]
    \centering
    \includegraphics[width=0.9\linewidth]{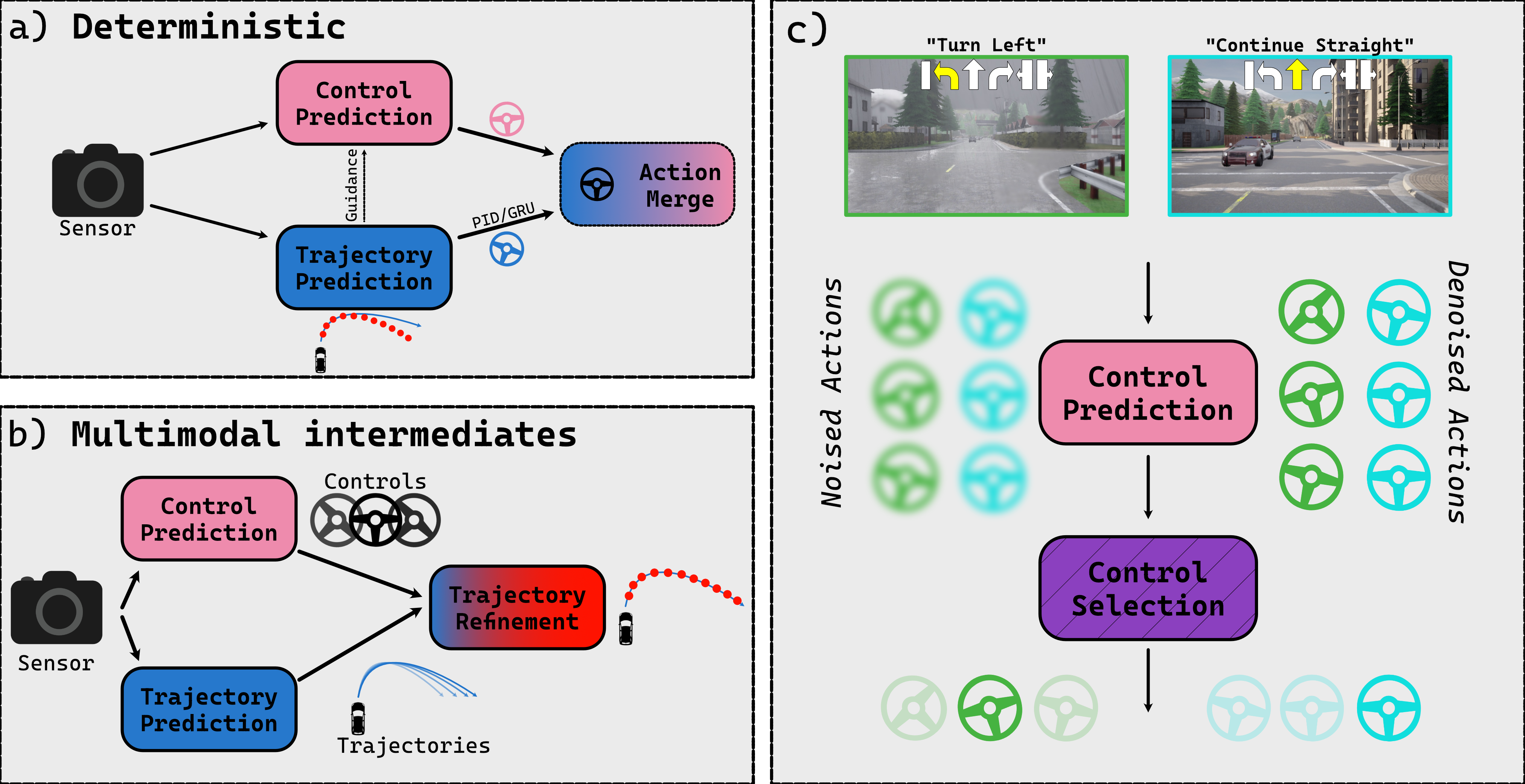}
    \caption{\textbf{Ambiguity in driving.} (a) \textbf{Deterministic control:} for a given scene,  we can predict the control to apply to the vehicle~\cite{Codevilla18,xiao2023cilpp}, plan its future trajectory~\cite{Chitta22TransFuser,Shao22InterFuser,Hamdan25ETA}, or fuse both branches before execution~\cite{Wu22TrajectoryGuided}. (b) \textbf{Multimodal intermediates:} multiple future trajectories can be generated or refined before being converted into control, but multimodality is expressed through intermediate trajectory representations ~\cite{Li24HydraMDP,Li25HydraNext}. (c) \textbf{Action multimodality (ours):} multiple control candidates are sampled directly in action space, and one candidate is selected for execution, enabling deployable multimodality without trajectory anchors or waypoint intermediates. Crucially, the candidate distribution reflects  the inherent ambiguity of the command: a \emph{Turn Left} instruction admits many valid steering profiles, producing a diverse candidate set, whereas \emph{Continue Straight} is geometrically constrained, yielding tighter agreement among candidates.}
    \label{fig:main_fig}
\end{figure*}

A key challenge in autonomous driving is that the future is fundamentally uncertain and multi-modal. At any decision point, multiple behaviours may be valid; for instance, a vehicle may continue straight, change lanes, or brake depending on the behaviour of surrounding agents, and committing prematurely to a single prediction risks catastrophic failure in safety-critical scenarios. This ambiguity is structural: it arises from partial observability and occlusion, from interactions with other agents, and from the inherent unpredictability of human behaviour. Representing a single deterministic future is therefore insufficient for robust decision-making.

In robotics, Diffusion Policy~\cite{Chi23DiffusionPolicy} models multimodality directly in action space by learning a conditional denoising process over action sequences, which are sampled and executed in a receding-horizon manner. Importantly, however, Diffusion Policy represents multimodality \emph{implicitly} through the learned action distribution: it does not explicitly expose a set of candidate actions at each decision step, nor does it rank or select among such candidates before execution. 

In autonomous driving, multimodality has instead been exploited primarily in trajectory space. Representative methods include DiffusionDrive~\cite{Liao24}, which samples trajectory hypotheses around learned anchors, and Hydra-based approaches, which generate diverse trajectory candidates via distillation ~\cite{Li24HydraMDP} or control-derived trajectory proposals ~\cite{Li25HydraNext}. As illustrated in Figure~\ref{fig:main_fig}, previous paradigms either suppress multimodality at the control interface~(a), or express it only through trajectory intermediates before collapsing to a single output~(b). Our approach~(c) is the first to maintain and exploit multimodality directly in action space, deferring commitment to a single control only at execution time.

Overall, existing E2E-AD systems rely on trajectory-centric formulations, while direct modelling of multimodal actions remains less explored. Direct-control policies align naturally with the closed-loop execution of real vehicles, where actions are recomputed at every control cycle from fresh observations, enabling rapid adaptation to dynamic environments. Crucially, they are \emph{localisation-free}: operating without GPS or HD maps and relying solely on onboard cameras and a high-level command. Unlike recent robotics policies that often use action chunking to predict and execute short sequences of future actions~\cite{Pertsch25FAST} to account for a large model latency, we adopt a single-step horizon to maximise reactivity in closed-loop driving, replanning the executed control at every timestep. In this work, we focus on \emph{direct, control-based} E2E-AD with explicit multimodal action prediction.

Inspired by Diffusion Policy~\cite{Chi23DiffusionPolicy}, we introduce \textbf{Action Diffusion Transformer (ADT)}, an anchor-free diffusion transformer that makes action-space multimodality explicit by generating multiple candidate actions and selecting the most suitable one at inference through Nearest Neighbour Matching (NNM).

In summary, our main contributions are:
\begin{itemize}
    \item \textbf{Action multimodality.} We demonstrate the value of action-space multimodality for direct control. Across offline and closed-loop evaluation, this strategy improves over deterministic control policies and naive stochastic baselines, demonstrating the value of multimodal action prediction for robust control-based E2E-AD.
    \item \textbf{A diffusion-based action-space formulation for direct-control E2E-AD.} ADT operates purely in action space, avoiding any reliance on trajectory representations or GPS-dependent controllers. Diffusion enables ADT to generate diverse action candidates rather than a single deterministic output, and NNM provides a lightweight, parallelisable selection mechanism that preserves low latency.
\end{itemize}

\section{Related Work}
\label{sec:related work}

\textbf{Reinforcement learning and imitation learning frameworks.} E2E-AD is commonly framed under two learning paradigms: Reinforcement Learning (RL) and Imitation Learning (IL). RL learns through trial-and-error interaction with the environment, making it appealing for optimising long-horizon driving behaviour. However, in real-world driving, unsafe exploration is infeasible, reward design is difficult, and privileged ground-truth state is not available to the deployed agent. In simulation, by contrast, such privileged information can be accessed and unsafe actions can be executed without physical risk, making RL particularly useful for training strong teacher policies and expert planners~\cite{Zhang21RLICoach,Li24Think2Drive}. IL instead learns directly from expert demonstrations with a supervised objective, avoiding online exploration and hand-crafted reward design. For these reasons, we adopt the IL framework and focus on direct action prediction from expert driving data.

\textbf{Closed-loop and open-loop paradigms.} Two complementary evaluation protocols shape the progress of E2E-AD. In \emph{open-loop} evaluation, the model output is scored against recorded expert futures, without feeding its predictions back to the environment~\cite{Dauner24NAVSIM}. This setting is scalable and efficient, which makes it attractive for large-scale development and benchmarking. However, its main limitation is that it ignores the feedback loop between the ego vehicle and its surroundings, and therefore does not fully reflect the interactive nature of real driving~\cite{Codevilla_2018_ECCV,Zhai23RethinkingOpenLoop,Li2023IsES}. In \emph{closed-loop} evaluation, by contrast, the policy is rolled out online so that each prediction affects future observations and the behavior of surrounding agents~\cite{Dosovitskiy17}. This makes the setting more faithful to the interactive nature of driving. 

\textbf{Trajectory-based and action-based approaches.} E2E-AD methods commonly predict either future trajectories or low-level control actions, with neither representation being universally preferable~\cite{Wu22TrajectoryGuided}. Trajectory-based methods reason over a longer future horizon and provide spatial intermediate goals that can facilitate human inspection, verification, and integration with downstream prediction and planning modules~\cite{Zablocki22Explainability,Wu22TrajectoryGuided,Jiang26VADv2,Liu26GuideFlow}. However, these trajectories must subsequently be converted into executable controls, typically using PID or more sophisticated controllers, and reliable trajectory following is itself a non-trivial problem~\cite{Wu22TrajectoryGuided}. Action-based methods instead optimise steering, throttle, and brake directly, avoiding this trajectory-to-control conversion and favouring reactive, potentially localisation-free closed-loop operation~\cite{Wu22TrajectoryGuided}. Thus, trajectory-space and action-space formulations optimise for different objectives: the former favour interpretable and verifiable long-horizon planning, whereas the latter favour direct and reactive execution. The preferable formulation therefore depends on the target application and available data; in this work, we focus on direct action prediction from expert control demonstrations.

\textbf{Conditional imitation and action-based models.}
Early deep E2E systems lacked route-level intent, learning how to drive locally but not which manoeuvre to take at intersections. Conditional Imitation Learning (CIL) addressed this by conditioning control prediction on high-level commands such as turning left or going straight~\cite{Codevilla18}. Hawke \emph{et al.} extended CIL to real-world urban driving, training from human demonstrations to jointly control steering and speed over unseen routes~\cite{Hawke20UrbanCIL}. CILRS~\cite{Codevilla19BehaviorCloning} further studied large-scale behaviour cloning in CARLA~\cite{Dosovitskiy17}, exposing dataset bias, generalisation gaps, and closed-loop instability in action-based policies. More recent control-based models include CIL++, which improves vision-only control with high-resolution multi-view inputs and attention~\cite{xiao2023cilpp}, MILE, which learns a camera-only world model jointly with a control policy~\cite{Hu22MILE}, and D$^3$Nav, which couples compact latent prediction to driving in unstructured traffic~\cite{Ganesh24D3Nav}. These methods differ from recent work that focuses on richer intermediate representations, such as tracking-aware driving LMMs~\cite{Ishaq25TrackingMeetsLMM}, probabilistic BEV mapping~\cite{Erdogan25MappingLikeASkeptic}, and BEV distillation for camera-only perception~\cite{Kim25DualDistill}. Since CIL++ and MILE predict low-level controls directly from cameras without privileged maps or LiDAR, they are natural vision-only baselines for closed-loop comparison.

\textbf{Generative modelling and multimodality.}
Recent generative approaches in autonomous driving mostly express multimodality through intermediate planning representations. Beyond DiffusionDrive~\cite{Liao24}, recent work has further explored generative modelling for trajectory refinement and multi-candidate planning. BridgeDrive, DiffRefiner and DiffE2E extend this direction by refining coarse or proposed trajectories through diffusion-based planning modules~\cite{Liu26BridgeDrive,Yin26DiffRefiner,Zhao25DiffE2E}. Beyond diffusion, Hydra-MDP learns diverse trajectory candidates through multi-target Hydra-distillation~\cite{Li24HydraMDP}. Hydra-NeXt produces multiple diffusion-based control proposals, but these are rolled out with a Kinematic Bicycle Model (KBM) and matched in trajectory space~\cite{Li25HydraNext}. Generative modelling has also been used for driving data and sensor realism, including LiDAR object generation with point diffusion~\cite{Kirby25LOGen} and labelled LiDAR synthesis with neural radiance fields~\cite{Srivastava25LidLabNeRF}. In contrast, ADT makes multimodality explicit in continuous action space by generating throttle, steer, and brake candidates and selecting among them directly with Nearest Neighbour Matching (NNM).  


\section{Method}
\label{sec:method}

\subsection{Preliminaries}
\label{sec:preliminaries}

\paragraph{Problem formulation.}
We consider end-to-end conditional control from multi-view visual observations and driving measurements. At control timestep $t$, the policy receives the observation
\begin{equation}
    o_t
    =
    \left(
        \left\{ I^{c} \right\}_{c=1}^{C},
        \mathbf{g}_t,
        v_t
    \right),
    \label{eq:observation}
\end{equation}
where $I^{c}$ denotes the RGB observation from camera $c$, $\mathbf{g}_t \in \{0,1\}^{N_{\mathrm{cmd}}}$ is a one-hot high-level navigation command, and $v_t \in \mathbb{R}$ is the current ego-vehicle speed.

Given a dataset of expert demonstrations $\mathcal{D}=\left\{\left(o_n, \mathbf{u}_n\right)\right\}_{n=1}^{N}$, where $\mathbf{u}_n$ denotes the expert controls corresponding to the observations $o_n$, the goal is to learn a conditional action distribution $\pi_{\theta}\!\left(\mathbf{u}_t \mid o_t\right)$, rather than a deterministic regressor. This distinction is important in driving because a given observation may admit several plausible expert actions, for example, when different steering or speed-control decisions remain compatible with the same scene and command. Our model, therefore, represents the action distribution through conditional diffusion in control space.

\paragraph{Diffusion formulation.}
Let $\mathbf{x}_t^{0} := \mathbf{u}_t$ denote a clean expert control and let $\mathbf{x}_t^{r}$ denote its noisy version at diffusion stage $r \in \{1,\ldots,R\}$. A forward diffusion process progressively perturbs the clean action according to
\begin{equation}
    q\!\left(
        \mathbf{x}_t^{r}
        \mid
        \mathbf{x}_t^{r-1}
    \right)
    =
    \mathcal{N}
    \left(
        \mathbf{x}_t^{r};
        \sqrt{\alpha_r}\,\mathbf{x}_t^{r-1},
        \beta_r \mathbf{I}_3
    \right),
    \qquad
    \alpha_r := 1-\beta_r ,
    \label{eq:forward_markov}
\end{equation}
where $\{\beta_r\}_{r=1}^{R}$ is a prescribed variance schedule. 

The learned reverse process is conditioned on the observation:
\begin{equation}
    p_{\theta}
    \left(
        \mathbf{x}_t^{0:R}
        \mid
        o_t
    \right)
    =
    p\!\left(\mathbf{x}_t^{R}\right)
    \prod_{r=1}^{R}
    p_{\theta}
    \left(
        \mathbf{x}_t^{r-1}
        \mid
        \mathbf{x}_t^{r},
        o_t
    \right),
    \qquad
    p\!\left(\mathbf{x}_t^{R}\right)
    =
    \mathcal{N}\!\left(\mathbf{0},\mathbf{I}_3\right).
    \label{eq:conditional_reverse_process}
\end{equation}

Following denoising diffusion probabilistic models~\cite{Ho20DDPM,Song21DDIM,Chi23DiffusionPolicy}, we parametrise the reverse process through a neural network that predicts the Gaussian perturbation present in a noisy action.

\subsection{Architecture}
\label{sec:conditional_action_diffusion}

\paragraph{Observation-conditioned denoiser.}
Our Action Diffusion Transformer (ADT) first maps the observation $o_t$ to a compact sequence of observation-conditioning tokens. Visual features are extracted from all camera views and all input timesteps, then flattened into a token sequence:
\begin{equation}
    \mathbf{F}_t
    =
    E_{\mathrm{vis}}
    \left(
        \left\{ I^{c} \right\}_{c=1}^{C}
    \right)
    \in
    \mathbb{R}^{L \times d},
    \qquad
    L = C H' W',
    \label{eq:visual_tokens}
\end{equation}
where $H' \times W'$ is the spatial resolution of the visual feature map and $d$ is the token dimension. The command and speed are embedded into the same latent dimension and added to every visual token:
\begin{equation}
    \mathbf{X}_t
    =
    \mathbf{F}_t
    +
    \mathbf{1}_{L}
    e_{\mathrm{cmd}}\!\left(\mathbf{g}_t\right)^{\top}
    +
    \mathbf{1}_{L}
    e_{\mathrm{spd}}\!\left(v_t\right)^{\top}
    +
    \mathbf{P}_{\mathrm{obs}},
    \label{eq:measurement_conditioning}
\end{equation}
where $e_{\mathrm{cmd}}(\cdot),e_{\mathrm{spd}}(\cdot)\in\mathbb{R}^{d}$ are learned embeddings and $\mathbf{P}_{\mathrm{obs}}\in\mathbb{R}^{L\times d}$ is a positional embedding over the token sequence. A Transformer Encoder~\cite{Vaswani17} then contextualises the full set of visual and measurement-conditioned tokens:
\begin{equation}
    \mathbf{M}_t
    =
    \operatorname{TxEnc}_{\theta}
    \left(
        \mathbf{X}_t
    \right)
    \in
    \mathbb{R}^{L \times d}.
    \label{eq:observation_memory}
\end{equation}

Rather than passing all $L$ encoder tokens to the diffusion decoder, ADT extracts a short observation-token sequence using learned queries. Let $\mathbf{Q}_{o} \in \mathbb{R}^{T_o \times d}$ denote the learned observation queries, where $T_o$ is the number of observation tokens exposed to the diffusion denoiser. The compressed observation representation is
\begin{equation}
    \mathbf{Z}_t
    =
    \operatorname{MHA}_{\theta}
    \left(
        \mathbf{Q}_{o},
        \mathbf{M}_t,
        \mathbf{M}_t
    \right)
    \in
    \mathbb{R}^{T_o \times d}.
    \label{eq:observation_tokens}
\end{equation}

The diffusion denoiser receives a noisy action $\mathbf{x}_t^{r}$, the diffusion-stage embedding $\gamma(r)$, and the observation tokens $\mathbf{Z}_t$. Since our policy predicts the immediate low-level control, the action horizon is one and $\mathbf{x}_t^{r}\in\mathbb{R}^{3}$. The decoder memory is formed by concatenating the diffusion-stage token with the $T_o$ observation tokens:
\begin{equation}
    \mathbf{C}_t^{r}
    =
    \begin{bmatrix}
        \gamma(r) \\
        \mathbf{Z}_t
    \end{bmatrix}
    +
    \mathbf{P}_{\mathrm{cond}}
    \in
    \mathbb{R}^{(1+T_o) \times d},
    \label{eq:diffusion_condition_memory}
\end{equation}
where $\mathbf{P}_{\mathrm{cond}}$ is a learned positional embedding.

The noisy action is projected into the decoder feature space and decoded against this conditional memory:
\begin{equation}
    \mathbf{h}_t^{r}
    =
    \operatorname{TxDec}_{\theta}
    \left(
        e_{\mathrm{act}}\!\left(\mathbf{x}_t^{r}\right)
        +
        \mathbf{p}_{\mathrm{act}},
        \mathbf{C}_t^{r}
    \right),
    \label{eq:conditional_decoder}
\end{equation}
where $e_{\mathrm{act}}(\cdot)$ is a learned action embedding and $\mathbf{p}_{\mathrm{act}}$ is its positional embedding. We adapt the standard noise-prediction setting in diffusion models. Hence, the predicted noise is
\begin{equation}
    \widehat{\boldsymbol{\epsilon}}_{\theta}
    \left(
        \mathbf{x}_t^{r},
        r,
        o_t
    \right)
    =
    W_{\epsilon}
    \operatorname{LN}
    \left(
        \mathbf{h}_t^{r}
    \right)
    \in
    \mathbb{R}^{3}.
    \label{eq:noise_predictor}
\end{equation}

\paragraph{Action diffusion objective.}
During training, a clean demonstrated control $\mathbf{u}_t$ is perturbed at a randomly sampled diffusion stage $r$, and the network is trained to recover the injected noise. Specifically,
\begin{equation}
    r \sim \operatorname{Unif}\{1,\ldots,R\},
    \qquad
    \boldsymbol{\epsilon}
    \sim
    \mathcal{N}\!\left(\mathbf{0},\mathbf{I}_3\right),
    \qquad
    \mathbf{x}_t^{r}
    =
    \sqrt{\bar{\alpha}_r}\,\mathbf{u}_t
    +
    \sqrt{1-\bar{\alpha}_r}\,
    \boldsymbol{\epsilon}.
    \label{eq:training_corruption}
\end{equation}
The denoising objective is the mean squared error between the sampled noise and the predicted noise:
\begin{equation}
    \mathcal{L}_{\mathrm{diff}}(\theta)
    =
    \mathbb{E}_{(\mathbf{u}_t,o_t)\sim\mathcal{D},\,r,\,\boldsymbol{\epsilon}}
    \left[
        \frac{1}{3}
        \left\|
            \boldsymbol{\epsilon}
            -
            \widehat{\boldsymbol{\epsilon}}_{\theta}
            \left(
                \mathbf{x}_t^{r},
                r,
                o_t
            \right)
        \right\|_2^2
    \right].
    \label{eq:diffusion_loss}
\end{equation}
Thus, the model learns a conditional distribution over the next continuous vehicle control rather than regressing to a single conditional mean. The one-step formulation is consistent with closed-loop driving: after executing the selected control, the policy observes the next scene and samples a new control conditioned on the updated observation. Our model samples actions at inference via DDIM. Further details on DDIM action sampling can be found in the supplementary material.

\paragraph{Nearest Neighbour Matching.}
Although conditional diffusion can represent multiple plausible controls, sampling a single reverse trajectory exposes only one random realisation of that distribution. We make the multimodality explicit at inference time by drawing $K>1$ initial Gaussian states under the same observation condition:
\begin{equation}
    \boldsymbol{\xi}^{(k)}
    \sim
    \mathcal{N}\!\left(\mathbf{0},\mathbf{I}_3\right),
    \qquad
    \widehat{\mathbf{u}}_t^{(k)}
    =
    \operatorname{DDIM}_{\theta}
    \left(
        \boldsymbol{\xi}^{(k)};
        \mathbf{Z}_t
    \right),
    \qquad
    k=1,\ldots,K.
    \label{eq:multi_candidate_sampling}
\end{equation}
All candidates are generated by the same trained denoiser and share the same observation-token sequence $\mathbf{Z}_t$; they differ only through their diffusion initialisations.

Nearest Neighbor Matching (NNM) selects one of these sampled controls by measuring candidate agreement in an execution-compatible action space. Let
\begin{equation}
    \widetilde{\mathbf{u}}_t^{(k)}
    =
    \varphi
    \left(
        \widehat{\mathbf{u}}_t^{(k)}
    \right)
    \label{eq:matching_transform}
\end{equation}
denote the deterministic mapping from a raw diffusion output to its sanitised control representation, where $\varphi(\cdot)$ is also applied before actuation. The distance between two candidates is defined as the mean component-wise $L_1$ distance
\begin{equation}
    d_{k\ell}
    =
    \frac{1}{3}
    \left\|
        \widetilde{\mathbf{u}}_t^{(k)}
        -
        \widetilde{\mathbf{u}}_t^{(\ell)}
    \right\|_1,
    \qquad
    k,\ell \in \{1,\ldots,K\}.
    \label{eq:nnm_pairwise_distance}
\end{equation}
Each candidate receives a consensus score equal to its mean distance to the remaining samples:
\begin{equation}
    s_k
    =
    \frac{1}{K-1}
    \sum_{\substack{\ell=1 \\ \ell \neq k}}^{K}
    d_{k\ell}.
    \label{eq:nnm_consensus_score}
\end{equation}
The selected control is the candidate medoid,
\begin{equation}
    k^{\star}
    =
    \underset{k \in \{1,\ldots,K\}}{\arg\min}
    \; s_k,
    \qquad
    \mathbf{u}_t^{\mathrm{NNM}}
    =
    \widetilde{\mathbf{u}}_t^{(k^{\star})}.
    \label{eq:nnm_selection}
\end{equation}

\section{Experiments}
\label{sec:experiments}

\subsection{Dataset and Benchmark}

We evaluate Action Diffusion Transformer (ADT) on the challenging Bench2Drive benchmark~\cite{Jia24Bench2Drive}. We follow the standard Bench2Drive protocol and train on the official \emph{base} split, which contains $1{,}000$ clips (comparable in scale to nuScenes \cite{Caesar20nuScenes}). At evaluation time, we use the official \texttt{Bench2Drive220} and report the standard metrics: Driving Score (DS), Success Rate (SR), Efficiency (E), Comfortness (C), and the five multi-ability scores (Merging, Overtaking, Emergency Brake, Give Way, Traffic Sign), as in~\cite{Jia24Bench2Drive}. Moreover, we use the \texttt{Dev10} validation set~\cite{Jia25DriveTransformer} exclusively as a validation split for model development, while \texttt{Bench2Drive220} is the only test set used for final evaluation in our work. Further details can be found in the supplementary material. 

\subsection{Implementation Details}

ADT is implemented in PyTorch~\cite{Paszke_PyTorch_2019} with PyTorch Lightning~\cite{Falcon_PyTorch_Lightning_2019}. The visual backbone is a ResNet-34 encoder~\cite{He_Resnet_2016}, following CIL++~\cite{xiao2023cilpp}. ADT uses a sensor setup of two RGB cameras, facing the front and rear of the vehicle, together with the ego speed and high-level navigation command. We adopt such camera setup following Hydra-NeXt~\cite{Li25HydraNext}. We set the observational token $T_o=1$ and horizon $H=1$. Further details can be found in the supplementary material. 

\begin{table}[t]
\begin{center}
\resizebox{\linewidth}{!}{%
\begin{tabular}{llccccccc}
\toprule
\multicolumn{2}{c}{\textbf{Configuration}} &
\multicolumn{3}{c}{\textbf{Architectural properties}} &
\multicolumn{4}{c}{\textbf{\texttt{Bench2Drive220}} (Closed-loop)} \\
\cmidrule(lr){1-2}
\cmidrule(lr){3-5}
\cmidrule(lr){6-9}
\textbf{ID} &
\textbf{Variant} &
\shortstack{\textbf{Action Head}} &
\shortstack{\textbf{Action Selection}} &
\shortstack{\textbf{Multimodal}} &
$\textbf{DS}\uparrow$ &
$\textbf{E}\uparrow$ &
$\textbf{C}\uparrow$ &
$\textbf{MA}(\%)\uparrow$ \\
\midrule

A & CIL++~\cite{xiao2023cilpp} &
MLP $\rightarrow \hat{u}$  &
Deterministic &
\xmark &
$59.53$ &
$204.23$ &
$18.38$ &
$30.95$ \\

B & w/ Transf. Decoder &
Trf. Dec. $\rightarrow$ MLP $\rightarrow \hat{u}$ &
Deterministic &
\xmark &
$67.45$ &
$192.43$ &
$22.83$ &
$45.83$ \\

C & w/ Diffusion &
Trf. Dec. $\rightarrow \hat{\epsilon}$ &
Candidate 0 &
\cmark &
$70.14$ &
$188.98$ &
$24.76$ &
$49.13$ \\

D & w/ NNM &
Trf. Dec. $\rightarrow \hat{\epsilon}$ &
NNM &
\cmark &
$77.90$ &
$192.15$ &
$25.63$ &
$55.47$ \\

\bottomrule
\end{tabular}%
}
\end{center}
\caption{\textbf{Progressive architectural improvements towards ADT.}
All variants use the same two-camera setting. The architectural-property columns highlight changes not fully captured by the variant names: the original CIL++ head predicts controls from mean-pooled encoder tokens; variants A and B use an MLP action regressor; and NNM adds a parameter-free inference-time selector over sampled diffusion candidates. DS denotes Driving Score, E Efficiency, C Comfortness and MA the mean of the five ability scores. $\hat{u}$ denotes predicted actions and $\hat{\epsilon}$ predicted noise.}
\label{tab:ablation_architecture}
\end{table}

\subsection{Roadmap}
\label{sec:roadmap}

Our roadmap mirrors the progression from the MLP-based direct control regression of CIL++ towards ADT, and is summarised in Tab.~\ref{tab:ablation_architecture}. Configurations A--C are presented as design-motivating steps towards ADT (i.e. Config D), our proposed method. Further details on the roadmap can be found in the supplementary material. 

\paragraph{Config A: CIL++ baseline.}
We use CIL++~\cite{xiao2023cilpp} as the starting point. Multi-view image features are encoded into scene tokens, enriched with command and speed embeddings, and processed by a Transformer encoder. The resulting token sequence is globally averaged over the token dimension and passed to an MLP that directly regresses the low-level control $\hat{u}=(\hat{u}^{\mathrm{thr}},\hat{u}^{\mathrm{steer}},\hat{u}^{\mathrm{brk}})$. This produces a deterministic action for each observation.

\paragraph{Config B: Transformer decoder.}
We build on Config A by replacing the global-average-pooled action readout with a Transformer decoder~\cite{Vaswani17}. Following the query-based decoding principle used in driving architectures such as InterFuser~\cite{Shao22InterFuser}, a learned action query cross-attends to the encoded scene tokens and the decoded feature is then mapped by an MLP to the predicted control $\hat{u}$. This keeps the policy deterministic, but gives the action head a structured mechanism for attending to the relevant parts of the encoded observation.

\paragraph{Config C: Diffusion.}
We build on Config B by replacing direct action regression with conditional action diffusion, inspired by Diffusion Policy~\cite{Chi23DiffusionPolicy}. The Transformer decoder is repurposed as a denoising action head: it receives a noisy continuous control token, the diffusion timestep, and the observation-conditioning token, and predicts the injected noise $\hat{\epsilon}$ rather than directly predicting $\hat{u}$. At inference, a single Gaussian initialisation is denoised with DDIM to obtain one executable control sample; in the notation of Tab.~\ref{tab:ablation_architecture}, this corresponds to selecting Candidate 0.

\paragraph{Config D: Nearest Neighbour Matching.}
We build on Config C with an inference-only selection rule. For each observation, the diffusion pipeline samples $K$ candidate controls instead of one, and NNM executes the candidate with the smallest average distance to the others. Thus, NNM turns diffusion multimodality into an explicit best-of-$K$ decision, without adding trainable parameters, auxiliary supervision, retraining or fine-tuning.



\begin{table}[t]
\begin{center}
\resizebox{\linewidth}{!}{%
\begin{tabular}{l|cc|cccc}
\hline
\multirow{2}{*}{\textbf{Method}} & \multirow{2}{*}{\textbf{Sensors}} &
\multirow{2}{*}{\textbf{Latency} (ms) $\downarrow$} &
\multicolumn{4}{c}{\texttt{Bench2Drive220}} \\
\cline{4-7}
&
&
&
$\textbf{DS}\uparrow$ &
$\textbf{SR} (\%)\uparrow$ &
$\textbf{E}\uparrow$ &
$\textbf{C}\uparrow$ \\
\hline

TCP-ctrl*~\cite{Wu22TrajectoryGuided} &
Front Camera &
$83$ &
$30.47$ &
$7.27$ &
$55.97$ &
$\mathbf{51.51}$ \\

MILE$^{\dagger}$~\cite{Hu22MILE} &
Front Camera & 
$44$ &
$39.80$ &
$5.45$ &
$67.59$ &
$25.23$ \\

CIL++~\cite{xiao2023cilpp} &
Front/Back Cameras &
$\mathbf{5.4}$ &
$59.53$ &
$25.57$ &
$\mathbf{204.23}$ &
$18.38$ \\

DriveTransformer-Large~\cite{Jia25DriveTransformer} &
6 Cameras &
$211.7$ &
$63.46$ &
$35.01$ &
$100.64$ &
$20.78$ \\

DriveAdapter~\cite{Jia23DriveAdapter} &
6 Cameras &
$931$ &
$64.22$ &
$33.08$ &
$70.22$ &
$16.01$ \\

ETA~\cite{Hamdan25ETA} &
Front Camera &
$50$ &
$69.53$ &
$38.64$ &
$184.51$ &
$\underline{28.43}$ \\

Hydra-NeXt~\cite{Li25HydraNext} &
Front/Back Cameras &
$528.3$ &
$73.86$ &
$\underline{50.00}$ &
$\underline{197.76}$ &
$20.68$ \\

\rowcolor{gray!20}\textbf{ADT} &
Front/Back Cameras &
$\underline{19.2}$ &
$\mathbf{77.90}$ &
$\mathbf{55.00}$ &
$192.15$ &
$25.63$ \\

\hline
\end{tabular}%
}
\end{center}
\caption{\textbf{Closed-loop results of E2E-AD methods on Bench2Drive-220 evaluation set.}
Latency is reported in milliseconds. * denotes expert feature distillation.
$\dagger$ denotes regression-only implementation. All reported models in this table were previously trained with the official base dataset. DS denotes Driving Score, SR denotes Success Rate, E denotes Efficiency, and C denotes Comfortness.}
\label{general_B2D}
\end{table}

\begin{table}[t]
\begin{center}
\resizebox{\linewidth}{!}{%
\begin{tabular}{l|ccccc|c}
\hline
\multirow{2}{*}{\textbf{Method}} &
\multicolumn{6}{c}{\textbf{Ability} (\%) $\uparrow$} \\
\cline{2-7}
&
Merging &
Overtaking &
Emergency Brake &
Give Way &
Traffic Sign &
\textbf{Mean} \\
\hline


TCP-ctrl*~\cite{Wu22TrajectoryGuided} &
10.29 &
4.44 &
10.00 &
10.00 &
6.45 &
8.23 \\

MILE$^{\dagger}$~\cite{Hu22MILE} &
5.00 &
0.00 &
3.33 &
10.00 &
25.79 &
8.82 \\

CIL++~\cite{xiao2023cilpp} &
$28.75$ &
$35.56$ &
$20.00$ &
$30.00$ &
$40.43$ &
$30.95$ \\

DriveTransformer-Large~\cite{Jia25DriveTransformer} &
$17.57$ &
$35.00$ &
$48.36$ &
$40.00$ &
$\underline{52.10}$ &
$38.60$ \\

DriveAdapter~\cite{Jia23DriveAdapter} &
$14.55$ &
$22.61$ &
$54.04$ &
$\underline{50.00}$ &
$50.45$ &
$38.33$ \\

ETA~\cite{Hamdan25ETA} &
$26.66$ &
$50.42$ &
$\underline{60.13}$ &
$\mathbf{80.00}$ &
$43.64$ &
$52.17$ \\

Hydra-NeXt~\cite{Li25HydraNext} &
$\underline{40.00}$ &
$\mathbf{64.44}$ &
$\mathbf{61.67}$ &
$\underline{50.00}$ &
$50.00$ &
$\underline{53.22}$ \\

\rowcolor{gray!20}\textbf{ADT} &
$\mathbf{50.00}$ &
$\underline{57.78}$ &
$\mathbf{61.67}$ &
$\underline{50.00}$ &
$\mathbf{57.89}$ &
$\mathbf{55.47}$ \\

\hline
\end{tabular}%
}
\end{center}
\caption{\textbf{Multi-ability results of E2E-AD methods on the closed-loop Bench2Drive-220 evaluation set.} All reported models in this table were previously trained with the official base dataset.
* denotes expert feature distillation. $\dagger$ denotes regression-only implementation.}
\label{multiability}
\end{table}

\subsection{Results on \texttt{Bench2Drive220}}
\label{subsec:quant_b2d220} 
As illustrated in Tab.~\ref{general_B2D}, ADT achieves the best closed-loop performance on Bench2Drive220, reaching $77.90$ DS and $55.00\%$ SR while running at $19.2$ ms. Compared with direct-control baselines, ADT substantially improves over TCP-ctrl ($30.47$ DS, $7.27\%$ SR), MILE ($39.80$ DS, $5.45\%$ SR) and CIL++ ($59.53$ DS, $25.57\%$ SR), while also raising the mean ability score from CIL++'s $30.95\%$ to $55.47\%$. ADT also outperforms waypoint-based methods, including DriveTransformer-Large ($63.46$ DS), DriveAdapter ($64.22$ DS), ETA ($69.53$ DS) and Hydra-NeXt ($73.86$ DS), despite using only front/back cameras and being much faster than the strongest trajectory baseline, Hydra-NeXt ($19.2$ ms vs. $528.3$ ms). Additional \texttt{Bench2Drive220} quantitative results exploring ADT with more sensors and higher-resolution inputs, together with qualitative ADT results showing induced trajectories across multiple rollouts, are provided in the supplementary material. In the ability benchmark (Tab.~\ref{multiability}), ADT obtains the highest mean score ($55.47\%$), with the best results in Merging and Traffic Sign, tied best performance in Emergency Brake, and competitive scores in Overtaking and Give Way. The only metrics where ADT is not best are Efficiency, led by CIL++, and Comfortness, led by TCP-ctrl.

\paragraph{Note on comfortness.} ADT obtains substantially lower comfortness than TCP-ctrl. This is partly due to how Bench2Drive defines its comfort metric, which marks a fixed 20-step segment as comfortable only if all samples satisfy strict bounds on acceleration, jerk, and yaw motion. Agents that move very slowly, get stuck, or remain nearly stationary can therefore accumulate comfortable segments while making little progress, explaining the high comfortness of TCP-ctrl despite its low success rate. For ADT, the lower comfortness is more likely caused by the opposite behaviour: the policy completes more routes, which can introduce sharper throttle, brake, or steering changes. More details can be found in the supplementary material.


\subsubsection{Multimodality results}
\label{sec:multimodality_results}

We evaluate multimodality in three stages. First, we isolate the candidate-selection problem offline, asking whether the sampled action set contains useful alternatives and whether a deployable selection rule can recover them. Second, we analyse whether diffusion produces a richer action-space candidate set than a deterministic policy with stochastic dropout. Third, we evaluate all deployable variants closed-loop using the \texttt{Dev10} set.

\paragraph{Selecting the best action candidate.}
We first study how to select one control command from a set of $K$ candidates. Besides the non-parametric nearest-neighbour matching (NNM) rule, we implement a trainable candidate selector in ADT. This selector scores candidate actions and selects one candidate from the diffusion-generated set. The implementation details of the learned selector, as well as additional information about our multimodality study, are provided in the supplementary material.

\begin{figure*}[t]
    \centering
    \begin{minipage}[b]{0.48\linewidth}
        \centering
        \includegraphics[width=\linewidth]{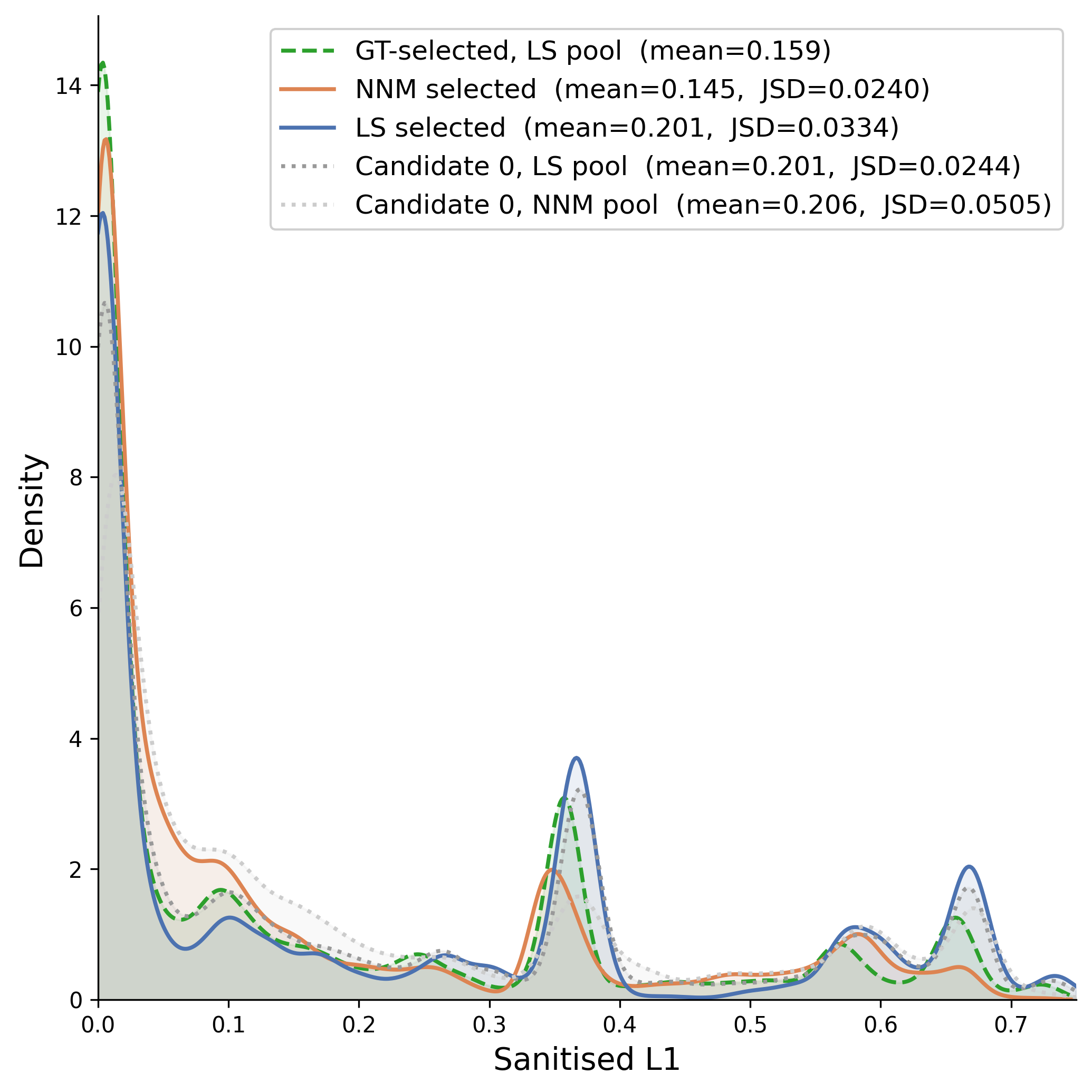}
        \vfill
    \end{minipage}
    \hfill
    \begin{minipage}[b]{0.45\linewidth}
        \centering
        \includegraphics[width=\linewidth]{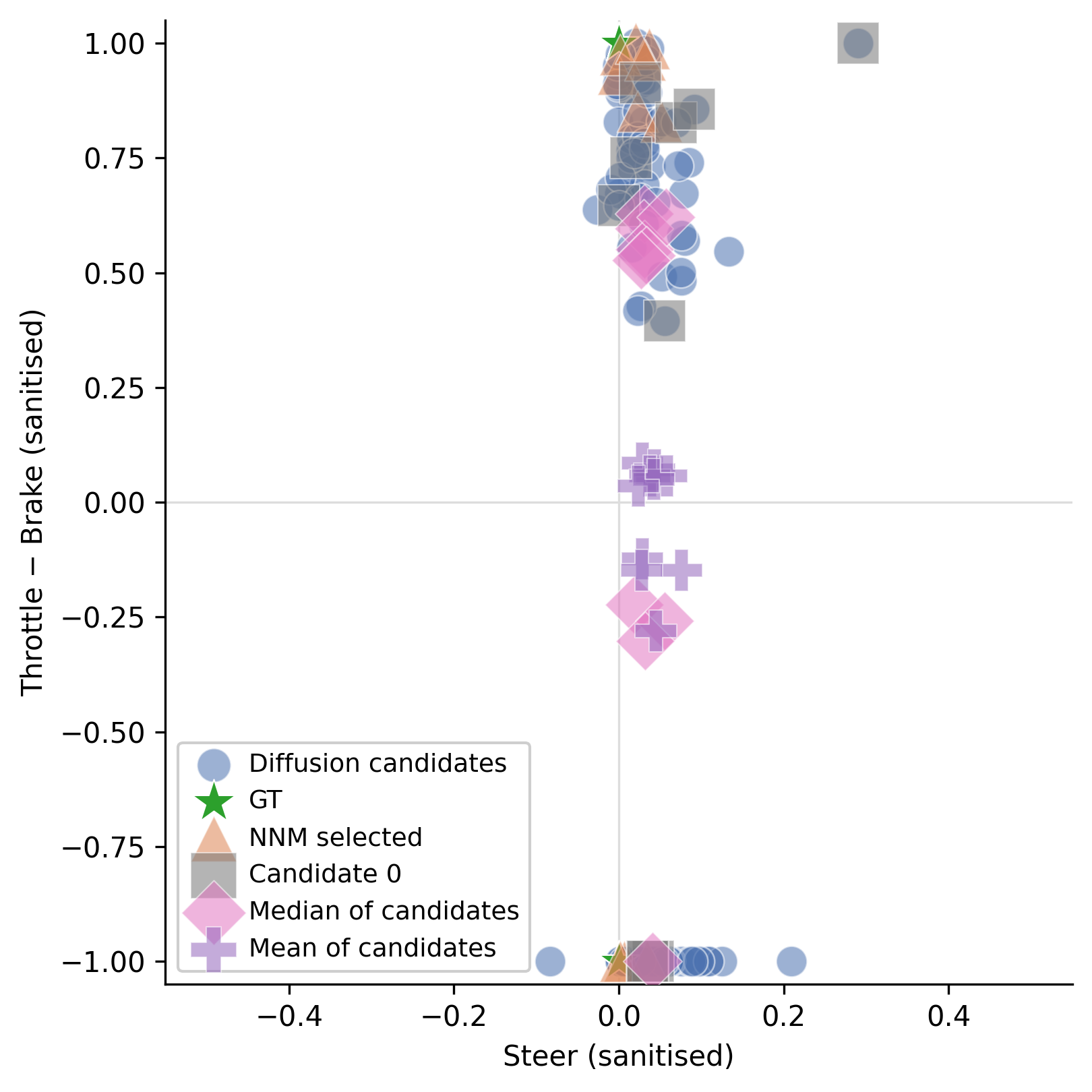}
    \end{minipage}

    \caption{\textbf{Offline candidate-selection.}
    \emph{Left:} We show the kernel density estimate of the total sanitised L1 error for each selection rule, measured against the GT-selected reference distribution. The legend reports the mean error and the Jensen--Shannon divergence (JSD) from the reference; lower values indicate better agreement. \emph{Right:} We place $10$ random action samples, each with $K=10$ candidates, in a lateral-longitudinal control grid. For each sample, we also plot their Candidate 0, median, and mean of the $K$ candidates.}
    \label{fig:offline_ls_vs_nnm}
\end{figure*}

\begin{table*}[t]
\begin{center}

\textbf{(a) Offline candidate-selection comparison}

\vspace{0.35em}

\resizebox{0.62\linewidth}{!}{%
\begin{tabular}{llcc}
\toprule
\textbf{Selection rule} &
Action space &
Selected L1 $\downarrow$ &
Oracle-best L1 $\downarrow$ \\
\midrule

\multirow{3}{*}{\textbf{NNM}}
& Raw       & \textbf{0.163} & \textbf{0.154} \\
& Clamped   & \textbf{0.159} & \textbf{0.151} \\
& Sanitised & \textbf{0.145} & \textbf{0.145} \\
\midrule

\multirow{3}{*}{Learned selector}
& Raw       & 0.174 & 0.167 \\
& Clamped   & 0.170 & 0.163 \\
& Sanitised & 0.159 & 0.159 \\

\bottomrule
\end{tabular}%
}

\vspace{1.0em}

\textbf{(b) Action-space diversity across $K=10$ candidates}

\vspace{0.35em}

\resizebox{0.86\linewidth}{!}{%
\begin{tabular}{lccccccc}
\toprule
\textbf{Model} &
\multicolumn{2}{c}{Throttle range $\uparrow$} &
\multicolumn{2}{c}{Steer range $\uparrow$} &
\multicolumn{2}{c}{Brake range $\uparrow$} &
Sanitised pairwise L1 $\uparrow$ \\
\cmidrule(lr){2-3}
\cmidrule(lr){4-5}
\cmidrule(lr){6-7}
\cmidrule(lr){8-8}
&
Raw & Sanitised &
Raw & Sanitised &
Raw & Sanitised &
Sanitised \\
\midrule

\textbf{ADT} &
\textbf{0.251} & \textbf{0.244} &
\textbf{0.123} & \textbf{0.123} &
\textbf{0.209} & \textbf{0.142} &
\textbf{0.060} \\

CIL++ w/ TD &
0.000 & 0.000 &
0.000 & 0.000 &
0.000 & 0.000 &
0.000 \\

CIL++ w/ TD w/ dropout &
0.193 & 0.175 &
0.014 & 0.014 &
0.045 & 0.008 &
0.024 \\

\bottomrule
\end{tabular}%
}

\end{center}

\caption{\textbf{Offline multimodality results.}
Panel (a) compares deployable candidate-selection rules. Selected denotes the candidate chosen by the corresponding rule, and Oracle-best denotes the candidate closest to the expert action offline. Panel (b) measures action-space diversity across $K=10$ candidates. ADT produces substantially larger and more balanced diversity than deterministic CIL++ w/ TD and CIL++ w/ TD with MC-dropout, while NNM obtains the lowest selected L1 among the evaluated selection rules.}
\label{tab:offline_multimodality_diagnostics}
\end{table*}

Figure~\ref{fig:offline_ls_vs_nnm} analyses candidate selection from two complementary views. On the left, NNM has the lowest total sanitised L1 and stays closest to the GT-selected reference distribution, while the learned selector leaves more mass at larger errors. On the right, we see that using the median or mean of the $K$ candidates leads us to \emph{coasting} actions, where there is no lateral or longitudinal motion. This happens because the sampled actions can split into throttle and brake modes, and averaging or taking the median falls between them. NNM avoids this by selecting an actual sampled action supported by nearby candidates.

Table~\ref{tab:offline_multimodality_diagnostics} (a) reports the corresponding aggregate errors under raw, clamped, and sanitised action spaces. Both candidate sets contain useful alternatives, but NNM recovers them more reliably, reaching a sanitised selected L1 of $0.145$ compared with $0.159$ for the learned selector. The oracle-best values quantify the remaining candidate-set potential, showing that multimodality must be paired with a robust deployable selection rule.

\paragraph{Diffusion multimodality.}
We next test whether diffusion provides a stronger mechanism for action-space multimodality than adding stochasticity to a deterministic policy. We start from the deterministic CIL++ w/ TD policy (see configuration B in Tab.~\ref{tab:ablation_architecture}). This policy has no explicit multimodal mechanism: for a fixed input, repeated forward passes produce the same action.

As a simple stochastic baseline, we add dropout to the policy MLP and keep dropout active at inference. Dropout was originally introduced as a regularisation method~\cite{Srivastava14Dropout}, and Monte Carlo dropout interprets repeated stochastic forward passes as approximate Bayesian inference over model parameters~\cite{Gal16DropoutBayesian}. For a fixed condition $\mathbf{c}$, this produces
\begin{equation}
    \hat{\mathbf{u}}^{(k)}
    =
    f_{\theta}\left(\mathbf{c}; \mathbf{m}^{(k)}\right),
    \qquad
    k=1,\ldots,K,
\end{equation}
where $\mathbf{m}^{(k)}$ is the dropout mask for sample $k$. This gives a stochastic candidate generator, but it does not define a learned conditional distribution over valid driving modes. In contrast, ADT samples $K$ action latents from a conditional diffusion model and denoises them into $K$ candidate controls.

Table~\ref{tab:offline_multimodality_diagnostics} (b) compares the resulting candidate spread for $K=10$. The deterministic CIL++ w/ TD model has zero diversity, as expected. MC-dropout introduces some variation: after sanitisation, its mean ranges are $0.175$ for throttle, $0.014$ for steer, and only $0.008$ for brake. ADT produces a broader and more balanced candidate set, with sanitised ranges of $0.244$, $0.123$, and $0.142$ for throttle, steer, and brake, respectively. The same trend appears in the sanitised pairwise L1 diversity, where ADT reaches $0.060$ compared with $0.024$ for MC-dropout and zero for the deterministic policy. This supports diffusion as a principled action-space multimodality mechanism rather than merely injecting local stochasticity into a deterministic action head.

\paragraph{Closed-loop multimodality.}
Finally, we evaluate whether offline multimodality translates into closed-loop driving performance. We compare several deployable ways of using the candidate set. For ADT, the vanilla variant uses a single sampled action, the mean variant averages the $K$ diffusion candidates, and NNM selects a candidate using nearest-neighbour matching in action space. For CIL++ w/ TD, we compare the deterministic policy, MC-dropout candidate 0, MC-dropout mean aggregation, and MC-dropout with NNM. The mean variants test whether simply generating multiple candidates is enough, or whether selecting one candidate is necessary.

NNM selects an action already sampled from the diffusion policy; it does not average potentially incompatible controls. More precisely, it favours a sampled control supported by nearby samples of the learned conditional action distribution, while isolated samples obtain larger consensus scores. This is important in closed loop: averaging can collapse distinct throttle, brake, or steering alternatives into a control that is not itself a likely candidate, while candidate selection preserves a valid sampled action.

\begin{table}[t]
\begin{center}
\resizebox{\linewidth}{!}{%
\begin{tabular}{lccccc}
\toprule
\textbf{Family} &
\textbf{Multimodality} &
\textbf{Variant} &
DS $\uparrow$ &
RC $\uparrow$ &
IP $\uparrow$ \\
\midrule

\multirow{3}{*}{\textbf{ADT}}
& \multirow{3}{*}{Diffusion}
& Candidate 0 &
$75.23^{\pm 1.86}$ &
$85.97^{\pm 2.31}$ &
$0.87^{\pm \mathbf{0.01}}$ \\

&
& Mean &
$78.32^{\pm 2.25}$ &
$89.77^{\pm \mathbf{1.36}}$ &
$\mathbf{0.89}^{\pm \mathbf{0.01}}$ \\

\rowcolor{gray!20}
&
& \textbf{NNM} &
$\mathbf{82.88}^{\pm \mathbf{1.83}}$ &
$\mathbf{93.12}^{\pm 1.91}$ &
$0.88^{\pm 0.02}$ \\

\midrule

CIL++ w/ TD
& \xmark 
& Deterministic &
$74.99^{\pm 2.50}$ &
$84.10^{\pm 2.06}$ &
$0.91^{\pm \mathbf{0.01}}$ \\

\midrule

\multirow{3}{*}{CIL++ w/ TD}
& \multirow{3}{*}{Dropout}
& Candidate 0 &
$59.13^{\pm 9.54}$ &
$72.58^{\pm 7.52}$ &
$0.61^{\pm 0.10}$ \\

&
& Mean &
$62.76^{\pm 4.04}$ &
$76.93^{\pm 3.52}$ &
$0.64^{\pm 0.03}$ \\

&
& NNM &
$71.24^{\pm 3.11}$ &
$84.28^{\pm 3.20}$ &
$0.78^{\pm 0.03}$ \\

\bottomrule
\end{tabular}%
}
\end{center}
\caption{\textbf{Closed-loop \texttt{Dev10} comparison of ADT, deterministic CIL++ w/ TD, and dropout variants.}
ADT variants compare different ways of exploiting diffusion-generated action candidates: a single sampled candidate, mean aggregation, and nearest-neighbour matching (NNM).
The dropout variants evaluate whether MC-dropout provides a comparable action-space multimodality mechanism when applied to the deterministic CIL++ w/ TD policy.
NNM gives the strongest driving score and route completion among all evaluated variants, showing that diffusion-generated candidates combined with candidate selection provide the most effective closed-loop control.
We evaluate each model three times with different seeds and report mean and standard deviation.}
\label{tab:closed_loop_multimodality}
\end{table}

Table~\ref{tab:closed_loop_multimodality} shows that selection is critical. ADT with NNM achieves the best closed-loop performance, with a driving score of $82.88 \pm 1.83$ and route completion of $93.12 \pm 1.91$. ADT mean aggregation improves over candidate 0, but remains below NNM, indicating that averaging candidates does not exploit the full candidate-set structure.

The dropout baselines show a different trend. MC-dropout candidate 0 and mean aggregation substantially degrade CIL++ w/ TD, and dropout NNM recovers part of the lost performance but still remains below the deterministic policy. Thus, adding dropout to a deterministic action head does not provide the same closed-loop benefit as diffusion-based candidate generation. Overall, the strongest result comes from combining diffusion-generated action candidates with a robust selection rule.


\begin{table*}[t]
\begin{center}
\small
\setlength{\tabcolsep}{4pt}

\resizebox{\linewidth}{!}{%
\begin{tabular}{lccccc|c|ccc}
\toprule
ID &
Diffusion &
Time Cond. &
Obs. Token &
Denoiser &
Selection &
Latency (ms) $\downarrow$ &
DS $\uparrow$ &
RC $\uparrow$ &
IP $\uparrow$ \\
\midrule
A1 & \xmark          & \xmark          & cond-query & TD  & \xmark          & \textbf{7.5} & 44.59 & 58.39 & 0.71 \\
A2 & \cmark  & \xmark          & cond-query & TD  & NNM         & 19.2 & 56.83 & 76.04 & 0.72 \\
A3 & \cmark  & \cmark  & mean       & TD  & NNM         & 19.2 & 62.78 & 91.23 & 0.71 \\
A4 & \cmark  & \cmark  & cond-query & MLP & NNM         & 18.5 & 69.72 & 84.98 & 0.81 \\
A5 & \cmark  & \cmark  & cond-query & TD  & Candidate 0 & 11.3 & 75.23 & 85.97 & 0.87 \\
\rowcolor{gray!20}\textbf{ADT} &
\cmark &
\cmark &
cond-query &
TD &
NNM & 19.2 & \textbf{82.88} & \textbf{93.12} & \textbf{0.88} \\
\bottomrule
\end{tabular}
}
\caption{\textbf{Ablations of the Action Diffusion Transformer in \texttt{Dev10}.}
We toggle the diffusion head, timestep conditioning, observation token, denoiser architecture, and candidate-selection strategy while keeping the perception backbone fixed. DS denotes Driving Score, RC denotes Route Completion, and IP denotes Infraction Penalty. We highlight our selected variant in grey and bold the best metrics.}
\label{tab:adt_arch_ablation}

\vspace{0.8em}

\begin{minipage}[t]{0.32\linewidth}
    \centering
    \resizebox{\linewidth}{!}{%
    \begin{tabular}{c|c|ccc}
    \toprule
    $n_d$ & Latency (ms) $\downarrow$ & DS $\uparrow$ & RC $\uparrow$ & IP $\uparrow$ \\
    \midrule
    1 & \textbf{10.6} & 74.97 & 85.77 & 0.87 \\
    \rowcolor{gray!20}2 & 19.2 & 82.88 & 93.12 & 0.88 \\
    4 & 48.1 & 82.94 & 93.41 & 0.86 \\
    6 & 68.8 & 83.33 & 93.98 & 0.88 \\
    8 & 86.2 & 84.57 & 94.11 & \textbf{0.89} \\
    10 & 102.6 & \textbf{85.41} & \textbf{95.01} & 0.88 \\
    \bottomrule
    \end{tabular}%
    }
    \captionof{table}{\textbf{Denoising steps $n_d$.}}
    \label{tab:adt_denoising_steps}
\end{minipage}
\hfill
\begin{minipage}[t]{0.32\linewidth}
    \centering
    \resizebox{\linewidth}{!}{%
    \begin{tabular}{c|c|ccc}
    \toprule
    $H$ & Latency (ms) $\downarrow$ & DS $\uparrow$ & RC $\uparrow$ & IP $\uparrow$ \\
    \midrule
    \rowcolor{gray!20}1 & \textbf{19.2} & \textbf{82.88} & \textbf{93.12} & 0.88 \\
    2 & \textbf{19.2} & 73.78 & 91.10 & 0.81 \\
    4 & \textbf{19.2} & 72.38 & 83.38 & \textbf{0.89} \\
    6 & \textbf{19.2} & 63.59 & 81.69 & 0.82 \\
    8 & \textbf{19.2} & 59.88 & 83.85 & 0.71 \\
    10 & \textbf{19.2} & 45.19 & 61.47 & 0.69 \\
    \bottomrule
    \end{tabular}%
    }
    \captionof{table}{\textbf{Horizon length $H$.}}
    \label{tab:adt_horizon_length}
\end{minipage}
\hfill
\begin{minipage}[t]{0.32\linewidth}
    \centering
    \resizebox{\linewidth}{!}{%
    \begin{tabular}{c|c|ccc}
    \toprule
    $K$ & \textbf{Latency} (ms) $\downarrow$ & \textbf{DS} $\uparrow$ & \textbf{RC} $\uparrow$ & \textbf{IP} $\uparrow$ \\
    \midrule
    1 & \textbf{19.2} & 74.93 & 85.15 & 0.87 \\
    2 & \textbf{19.2} & 75.68 & 85.78 & 0.87 \\
    4 & \textbf{19.2} & 77.14 & 87.08 & 0.87 \\
    6 & \textbf{19.2} & 78.97 & 88.51 & 0.87 \\
    8 & \textbf{19.2} & 81.67 & 91.89 & 0.87 \\
    \rowcolor{gray!20}10 & \textbf{19.2} & \textbf{82.88} & \textbf{93.12} & \textbf{0.88} \\
    12 & \textbf{19.2} & 81.97 & 92.11 & 0.87 \\
    \bottomrule
    \end{tabular}%
    }
    \captionof{table}{\textbf{Number of action candidates $K$.}}
    \label{tab:adt_candidate_number}
\end{minipage}

\end{center}
\end{table*}

\subsection{Ablation Study}
\label{sec:ablation}

\paragraph{Decoder and conditioning.}
Table~\ref{tab:adt_arch_ablation} studies the main architectural components of ADT. Removing diffusion yields the weakest model (A1), showing that deterministic single-action prediction is not sufficient in this setting. Adding diffusion without timestep conditioning improves performance (A2), and adding timestep conditioning further helps (A3), confirming that the denoising process benefits from explicit diffusion-time information. Replacing the mean observation token with a learned conditional query improves performance further (A3$\rightarrow$A4), while the Transformer-decoder denoiser gives a stronger action mapper than the MLP denoiser (A4$\rightarrow$ADT). Finally, comparing A5 with ADT isolates the effect of candidate selection: both use the same diffusion, timestep conditioning, cond-query token and TD denoiser, but NNM selection raises DS from $75.23$ to $82.88$. Overall, the best configuration combines time-aware diffusion, cond-query conditioning, a TD denoiser and NNM selection.

\paragraph{Denoising steps $n_d$.}
Table~\ref{tab:adt_denoising_steps} shows the latency--performance trade-off for the number of denoising steps. Moving from one to two steps gives the main gain, increasing DS from $74.97$ to $82.88$ while remaining below the $50$ ms budget required for $20$ FPS. Four steps stay just within this budget, but provide almost no additional DS. Beyond $n_d{=}4$, the improvements are small relative to the latency cost, and the model runs above $50$ ms, i.e., below $20$ Hz. We therefore use $n_d{=}2$.

\paragraph{Horizon length $H$.}
Table~\ref{tab:adt_horizon_length} evaluates action horizons larger than the default single-step control. In these ablations, the model predicts an $H$-step action sequence, but the agent replans at every simulator tick and executes only the first action of the newly predicted sequence. The best result is obtained with $H{=}1$, while longer horizons consistently reduce closed-loop performance. We also tested executing consecutive actions from the same predicted sequence before replanning, but this produced near-zero driving score. We therefore keep $H{=}1$. Effective multi-step action execution remains an open challenge, and exploring executable action sequences is therefore a promising direction for future work.

\paragraph{Candidate number $K$.}
Table~\ref{tab:adt_candidate_number} shows that increasing the number of sampled candidates steadily improves closed-loop performance up to $K{=}10$. This confirms that ADT benefits from sampling multiple plausible controls and selecting among them. However, increasing to $K{=}12$ does not improve over $K{=}10$, suggesting that the useful candidate coverage has saturated. Since the latency is effectively unchanged across this range in our batched implementation, we adopt $K{=}10$ as the default candidate count. Further ablations on ADT are provided in the supplementary material.




\section{Conclusion}
We introduced \emph{Action Diffusion Transformer} (ADT), an architecture that operates directly in action space by generating multiple throttle--steer--brake candidates and selecting the final command during inference with Nearest Neighbour Matching. Our experiments show that action-space multimodality is useful for end-to-end driving: diffusion produces richer and more balanced candidate sets than MC-dropout, while candidate selection is essential for translating multimodality into stronger closed-loop performance. Overall, ADT shows that direct control policies can benefit from explicit multimodal action prediction, and we hope this encourages further work on control-based E2E-AD for ADAS.

\section{Acknowledgements}

This research was conducted within the framework of the HAMILTON project (PID2024-157936NBI00), funded by MICIU/AEI/10.13039/501100011033 and by FEDER, EU. The authors also gratefully acknowledge fruitful discussions within the European project BERTHA (Grant Agreement No. 101076360), particularly regarding multiple driver categories, which informed aspects related to this research. BERTHA is funded by the European Union. However, views and opinions expressed are those of the authors only and do not necessarily reflect those of the European Union or the European Climate, Infrastructure and Environment Executive Agency (CINEA). Neither the European Union nor the granting authority can be held responsible for them. Antonio M. López acknowledges financial support for his general research activities from ICREA under the ICREA Academia Program. The authors also acknowledge the support of the Generalitat de Catalunya through the CERCA Program and its ACCIÓ Agency for CVC’s general activities.

\bibliography{egbib}
\end{document}